\documentclass[conference]{style/IEEEtran}

\usepackage{hyperref}
\usepackage{xspace}
\usepackage{lipsum}
\usepackage{graphicx}
\usepackage{enumitem}
\usepackage{algorithmic}
\usepackage[ruled,vlined]{algorithm2e}
\usepackage{color}
\usepackage{tcolorbox}
\usepackage{float}
\usepackage{mathtools}
\usepackage{subcaption}
\usepackage{amsmath}
\usepackage{cleveref}
\usepackage{authblk}
\usepackage{booktabs}
\usepackage{multirow}

\newcommand*{\ie}{{\em i.e.,}\xspace}
\newcommand*{\etal}{{\em et al.}\@\xspace}
\newcommand*{\eg}{{\em e.g.,}\@\xspace}

\usepackage{pifont}
\newcommand{\cmark}{\ding{51}}%
\newcommand{\xmark}{\ding{55}}%

\begin{document}

\title{Hierarchical Federated Learning with Privacy}
\author{Varun Chandrasekaran\IEEEauthorrefmark{2}, Suman Banerjee\IEEEauthorrefmark{2}, Diego Perino\IEEEauthorrefmark{3}, Nicolas Kourtellis\IEEEauthorrefmark{3}, \vspace*{0.15cm} \\ 
Telefonica Research\IEEEauthorrefmark{3}, University of Wisconsin-Madison\IEEEauthorrefmark{2} 
}

\maketitle

\begin{abstract}
Federated learning (FL), where data remains at the federated clients, and where only gradient updates are shared with a central aggregator, was assumed to be private. Recent work demonstrates that adversaries with gradient-level access can mount successful inference and reconstruction attacks. In such settings, differentially private (DP) learning is known to provide resilience. However, approaches used in the status quo (\ie central and local DP) introduce disparate utility vs. privacy trade-offs. In this work, we take the first step towards mitigating such trade-offs through {\em hierarchical FL (HFL)}. We demonstrate that by the introduction of a new intermediary level where calibrated DP noise can be added, better privacy vs. utility trade-offs can be obtained; we term this {\em hierarchical DP (HDP)}. Our experiments with 3 different datasets (commonly used as benchmarks for FL) suggest that HDP produces models as accurate as those obtained using central DP, where noise is added at a central aggregator. Such an approach also provides comparable benefit against inference adversaries as in the local DP case, where noise is added at the federated clients.
\end{abstract}
\section{Introduction}
\label{sec:intro}

Federated learning (FL) is a paradigm to enable distributed learning.
Federated clients locally learn, and share the gradient updates associated with local learning with a central aggregator which performs a {\em global model update}. Apart from enabling {\em scalability}, FL is assumed to provide additional privacy benefits~\cite{mcmahan2017communication}.The data used for training never leave the data owner (or the federated clients, in this case), as in traditional machine learning (ML) settings where a centralized party must have access to data from all participating entities. However, recent work~\cite{melis2019exploiting,boenisch2021curious,geiping2020inverting,yin2021see,zhu2019federated,pasquini2021eluding} has shown that such learning algorithms are susceptible to attacks where the {\em private} training data can be reconstructed by observing intermediary calculations of the learning algorithm.

Differential privacy~\cite{dwork2014algorithmic} ($DP$) is a formal construct that provides guarantees on {\em privacy leakage} of various mechanisms, including stochastic gradient descent~\cite{abadi2016deep,chaudhuri2011differentially,10.1145/3035918.3064047} which is commonly used for empirical risk minimization (a component of learning approaches such as FL). It achieves this by adding calibrated noise at various stages of the mechanism. In conventional settings (\ie not FL), it has been shown that DP learning algorithms alleviate the aforementioned attacks~\cite{song2021systematic,geyer2017differentially,naseri2020toward}. However, they introduce an inherent trade-off between protecting {\em privacy} of the data used for model training, and {\em utility} of the model learned, often measured through test accuracy.

In FL, until now, DP guarantees have been investigated at two levels. At the central level (\ie central aggregator), DP noise can be added to the aggregated gradient update~\cite{konevcny2016federated} (\ie central DP or CDP). However, this assumes that the clients {\em trust} the central aggregator to not perform (malicious) inference on their data. Alternatively, DP noise can be added at the client locally (\ie local DP or LDP)~\cite{kasiviswanathan2011can,wei2020federated,zhao2020local}. This assumes that the clients {\em do not} trust any actor in the entire FL ecosystem. While both these approaches provide safeguards against the aforementioned privacy adversary (under different trust models), they introduce disparate privacy vs. utility (typically measured through model performance) trade-offs. In this work, we aim to understand if a compromise can be reached through an intermediary approach, by introducing \emph{hierarchies} in the learning process, while still providing formal DP guarantees. We call this {\em hierarchical DP} (or HDP).

The hierarchies we envision are deviations from the status quo of FL: clients form, or belong to {\em zones}; updates from clients within a zone are aggregated by a {\em super-node} (an elected/chosen client within the zone), and the updates from zones (\ie super-nodes) are aggregated at the central aggregator. While such a hierarchy seems fabricated, observe that such hierarchies are omnipresent in daily computing systems (such as in telecommunication networks, the internet infrastructure etc.). One example could be the existence of compute-enabled routers which can act as super-nodes within a home environment (\ie {\em trust in one's peers}). Furthermore, the onset of edge computing has resulted in the development of extensive, compute-capable edge base stations~\cite{liu2016paradrop}, which can also act as super-nodes for all clients within a particular geographic zone (\eg {\em trust in members of the same region or country}). These ``naturally occurring'' clusters of clients (\eg~devices in the same household or office network, or gaming consoles joining and participating in the same P2P gaming server) have embedded trust in their formalization and incentives~\cite{marti2006taxonomy}.

In this paper, we take the first step in analyzing how hierarchies are beneficial with respect to privacy, and what are key technical challenges to be solved. The first challenge is in constructing such a hierarchy. We analyze different approaches that reflect the aforementioned scenarios, and discuss the trade-offs in \S~\ref{sec:supernode_selection}. The second challenge involves making modifications to the standard approaches to FL to ensure that the newly proposed HFL approach learns the same model as in the status quo. We discuss this in \S~\ref{sec:correctness}. The third challenge involves formalizing an algorithm (to provide DP) to be used in an HFL ecosystem. Determining the exact mechanism has direct implications on the privacy vs. utility trade-off, and provides avenues for {\em privacy amplification}, which we discuss in \S~\ref{sec:privacy}. We generalize the approach by considering scenarios of simultaneous needs for LDP, HDP and CDP in \S~\ref{sec:generalization}. Our final challenge is understanding the various threats faced by our new constructions. We first validate that the hierarchical construction, when combined with DP learning, is robust to reconstruction adversaries. We also analyze faulty behavior at an intra- and inter-zonal granularity, and describe the information leakage by adversaries in such settings, even when an adversarial client is elected as a super-node.

We highlight the contributions of our work below:
\begin{itemize}
\itemsep0em
\item We are the first to propose the notion of hierarchical DP, a consequence of adding noise in the newly proposed hierarchical FL setting.
\item We identify an opportunity for privacy amplification: the natural composability of the Gaussian mechanism provides more privacy at the central aggregator level when noise is added at the super-node.
\item We analyze prior attacks in FL setting, and comment on the applicability of the same in the context of hierarchical FL. In particular, we look into adversaries in super-node level and discuss potential adversarial inferences they may perform, such as data inference and reconstruction attacks. We show that hierarchical FL (with DP) can thwart such attacks.
\item We experimentally study the privacy vs. utility trade-off on different datasets and setups. We demonstrate that the (privacy and utility) performance of hierarchical FL sits between local and central DP. Surprisingly, we observe that the utility benefits obtained through our proposal are very close to that obtained by learning a model with central DP. 
\end{itemize}

\section{Background \& Related Work}
\label{sec:background}

We introduce different concepts used throughout the paper.

\vspace{1mm}
\noindent{\textbf{Federated Learning (FL)}} involves learning with many {\em federated clients} in a decentralized manner~\cite{konevcny2016federated}. At each {\em federated round}, the {\em central aggregator} shares its weights with all federated clients, each of which has a private dataset it is unwilling to divulge. At any given point in time, only $k \leq n$ clients may be {\em online}; the fraction $\frac{k}{n}$ is termed the {\em user selection probability}. Each client performs training for one epoch (\texttt{FedAvg}~\cite{mcmahan2017communication}) or multiple epochs (\texttt{FedSGD}~\cite{shokri2015privacy}) on its private dataset and shares the update (obtained by using stochastic gradient descent) with the central aggregator. The aggregator gathers all $k$ updates, and performs an aggregation operation on them. Then, the central aggregator updates its weights, and shares these updated weights to all federated clients to be used as the starting point for the next federated round. FL is assumed to provide privacy by design (since the data is not shared with any central aggregator). However, recent work has shown that by observing the client updates, an adversary may infer sensitive information about the private training data~\cite{melis2019exploiting,geiping2020inverting,boenisch2021curious,wang2019beyond,zhu2019federated,bhowmick2018protection,yin2021see}. Additionally, the convergence of the learning algorithm used by the central aggregator (and overall robustness) is subject to both (a) client availability~\cite{li2018federated, rajput2019detox, chen2018draco}, and (b) uniformity of the data distributed among the clients~\cite{zhao2018federated}. In this work, we do not focus on these two problems, but more so on the privacy problems associated with FL, and how they can be mitigated with differential privacy. 

\vspace{1mm}
\noindent{\textbf{Differential Privacy (DP)}} was proposed by Dwork \etal~\cite{dwork2014algorithmic}. Let $\varepsilon$ be a positive real number, and $\mathcal{A}$ be a randomized algorithm that takes a dataset as input. The algorithm $\mathcal{A}$ is said to provide  $\varepsilon$-DP if, for all datasets $D_1$ and $D_2$ that differ on a single element, and all subsets $S$ of the outcomes of running $\mathcal{A}$:  
\begin{equation*}
\Pr[{\mathcal {A}}(D_{1})\in S]\leq e^{\varepsilon}\cdot \Pr[{\mathcal {A}}(D_{2})\in S]
\end{equation*}
where the probability is over the randomness of the algorithm $\mathcal{A}$.
$\varepsilon$ is also known as the {\em privacy budget}. DP is achieved by adding {\em noise} chosen from a specific distribution to provide the aforementioned indistinguishability guarantee~\cite{geng2015staircase,balle2018improving}. 

\vspace{1mm}
\noindent{\textbf{Local \& Central DP:}} The traditional model defined earlier, also known as the \emph{central DP} ($CDP$) model, implicitly assumes the existence of a trusted entity that does not deviate from protocol specification and adds the calibrated noise to provide the DP guarantee. To alleviate these strong trust assumptions, \emph{local DP} ($LDP$)~\cite{kasiviswanathan2011can} assumes that each data contributor adds the noise locally, \ie in-situ. Such a mechanism has limited knowledge of the overall function being computed on all the data, and overestimates the amount of noise required to provide privacy. The relationship between LDP and CDP is dependent on the mechanism used to achieve DP. For example, the Laplacian mechanism~\cite{dwork2014algorithmic} ensures that $(\varepsilon,\delta)$-LDP also provides $(\varepsilon,\delta)$-CDP.

\vspace{1mm}
\noindent{\textbf{Private Machine Learning:}} In their seminal work, Chaudhuri \etal~\cite{chaudhuri2011differentially} perturb the outputs of empirical risk minimization (ERM) mechanisms. They proceed to introduce the notion of {\em objective perturbation}, which remains the cornerstone for DP learning. Abadi \etal~\cite{abadi2016deep} extend the notion of objective perturbation by proposing a DP variant of stochastic gradient descent, where noise is added to each gradient update calculated. They also provide a tight analysis of the privacy budget ($\varepsilon$) through the {\em moments accountant}. We point the curious reader to the work of Jayaraman \etal for more details on how DP learning is implemented in practice~\cite{jayaraman2019evaluating, zhao2020privacy-utility}. 
Orthogonal to the approaches involving DP are those that utilize techniques from cryptography, notable multi-party computation (or MPC)~\cite{gilad2016cryptonets,wagh2019securenn,zheng2019helen,mishra2020delphi} to provide data privacy.

\vspace{1mm}
\noindent{\textbf{FL with DP:}} McMahan \etal~\cite{mcmahan2017learning} propose the first approach where DP can be combined with FL to provide formal privacy guarantees. Similar ideas are proposed in the work of Geyer \etal~\cite{geyer2017differentially}. In both these settings, however, the central aggregator is able to observe either noise-free gradients or noisy gradients from corresponding clients. To break this connection, Bonawitz \etal~\cite{bonawitz2016practical,bonawitz2017practical} propose the notion of {\em secure aggregation}, a variant of MPC which provides the central aggregator an aggregated view of all gradients (noisy/non-noisy) from the clients. Truex \etal~\cite{truex2019hybrid} combine advances from LDP and MPC (through thresholding cryptography schemes) to provide a hybrid scheme to provide better privacy.  Truex \etal~\cite{truex2020ldp} also propose an approach using LDP. However, they formulate this based on an alternative DP definition.

\vspace{1mm}
\noindent{\bf Hierarchical FL (HFL):} Abad \etal~\cite{abad2020hierarchical} propose a mechanism to ensure communication-efficient and coordinated learning in the context of HFL.
Here, the notion of hierarchies stems from the presence of clients communicating with small base stations (or cellular towers) which act as intermediaries, who further communicate with macro base stations (or the central aggregator).
Similarly, Yuan \etal~\cite{yuan2020hierarchical} also propose a new protocol to optimize for communication efficiency in the LAN-WAN setting. This form of HFL is significantly different from that proposed by Briggs \etal~\cite{briggs2020federated} which aims to segregate clusters of {\em similar} clients which can be independently trained on heterogeneous models. Across these prior works, the actors are consistent: there are the federated clients as in the status quo. However, they interact with intermediary entities (such as base stations in the work of Abad \etal~\cite{abad2020hierarchical}), and these intermediary entities consolidate information (on the behest of a subset of the clients) for the central aggregator(s). It is important to note that prior work focuses on improving the scalability/communication-efficiency of FL through the introduction of hierarchies, whilst our contribution is studying the implications on privacy.

\begin{figure*}[t]
\centering
\includegraphics[width=1.5\columnwidth]{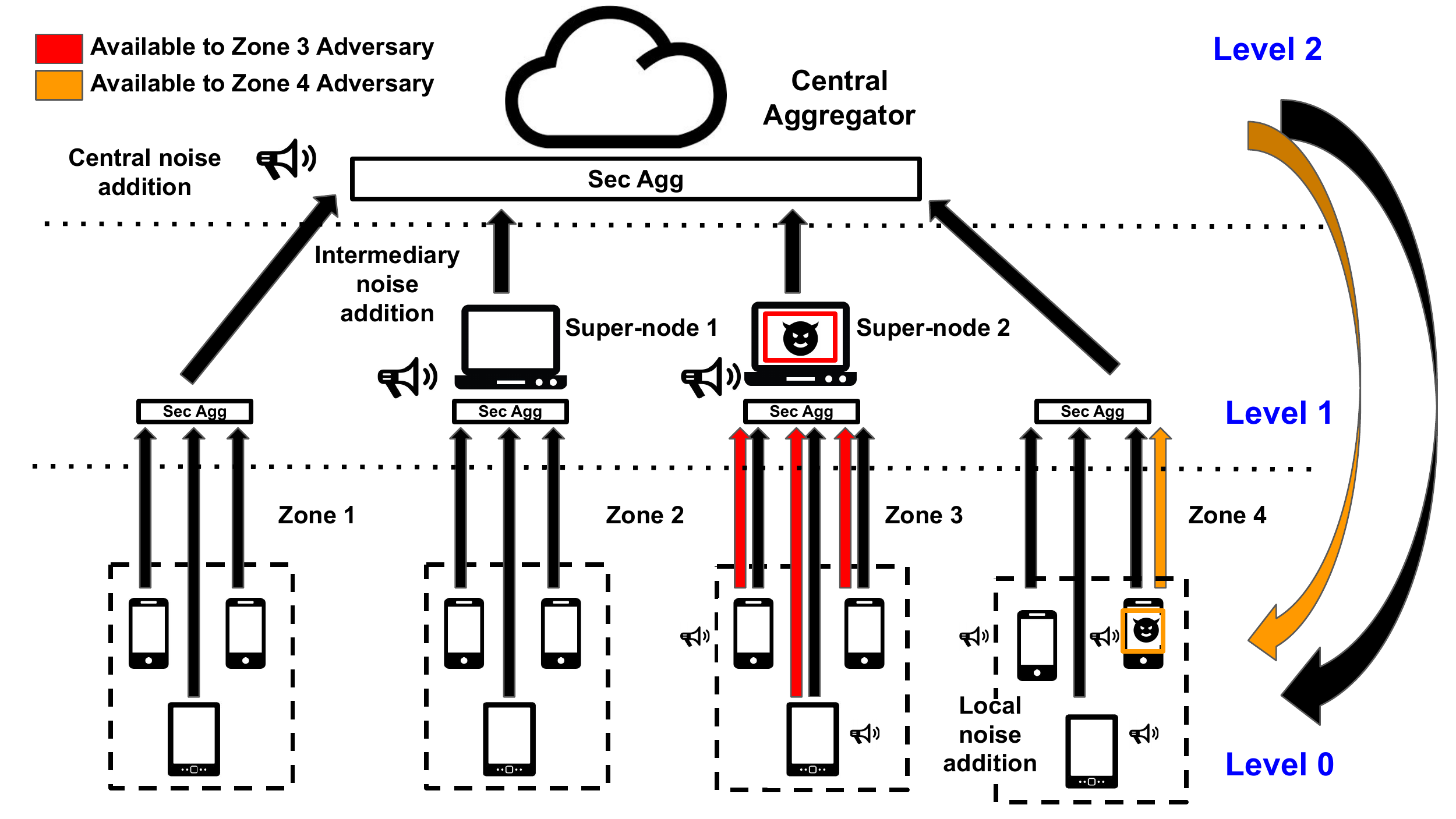}
\caption{An architectural overview of the approach. Different scenarios can be assumed to be present:
1) Clients in a zone that trust the central aggregator to add noise (CDP).
2) Clients in a zone that trust their intermediary super-node for noise addition (HDP).
3) Clients in a zone that do not trust anybody else and therefore add noise themselves (LDP).
4) Clients in a zone that may or not trust their intermediary super-node and chose (or not) to add noise (LDP+HDP).
For pure HDP, locally learnt gradients (from level 0) are perturbed with noise to provide formal DP guarantees at level 1.
This ensures that the gradients viewed by the central aggregator (at level 2) are less noisy, potentially resulting in a more performant model.
Adversarial nodes can be present either at level 0 (clients) or level 1 (super-node). 
}
\label{fig:architecture-overview}
\end{figure*}

\section{Approach Overview}
\label{sec:overview}

Recall from \S~\ref{sec:background} that FL requires $k \leq n$ online federated clients per round to collaborate with the central aggregator to jointly learn a model.  Thus, in the status quo, there is a total of two actors, at two conceptual levels of a hierarchy, \ie client(s) at level 0 and the server at level 1. In our proposal, we assume the existence of three main actors, at three different (conceptual) levels of the hierarchy (as noted in Figure~\ref{fig:architecture-overview}).

\vspace{1mm}
\noindent{\bf Explanation of Figure~\ref{fig:architecture-overview}:} At level 0 (\ie the lowest level), we have the federated clients who hold private data as in the status quo. At level 2 (\ie the highest level), we have the central aggregator, again as in the status quo. The key difference lies in the middle: at level 1 (\ie the intermediate level), we introduce a new entity called the {\em super-node}. The super-node is responsible for processing requests from the online clients in a particular region (or a {\em zone}). The presence of level 1 introduces a hierarchical approach to performing FL. An added feature in our proposal revolves around the selection of the super-node: they can either (a) be elected by a pool of their peers and thus are within the {\em same trust boundary/region} as their peers, or (b) be chosen as an entity in a different trust region (to both the clients and the central aggregator), but in geographic proximity to the clients chosen in the federated round.  Note that there can be many intermediary layers in practice, but for simplicity, we stick to one through the remainder of the paper. 

\vspace{1mm}
\noindent{\bf Impact on Privacy:} In the status quo, formal privacy guarantees in the FL setting are provided through the usage of DP.
In turn, this is obtained through two mechanisms: noise addition at the central aggregator (\ie CDP), or noise addition locally at each federated client (\ie LDP).
Empirical evidence suggests that the CDP mechanism will result in a final model with higher accuracy.
However, it makes a crucial assumption: the central aggregator (where the noise is added) is {\em trustworthy}.
The LDP mechanism removes this trust assumption at the expense of greater noise addition at each client.
To provide the best of both worlds, we propose {\em hierarchical DP} (or HDP).
Based on the construction described above, clients aggregate their (zonal-level) updates through the super-nodes, where calibrated noise is added, to provide the desired DP guarantee.
Through the remainder of the paper, we will describe modifications made to achieve HDP.

\begin{algorithm}[h]
\begin{algorithmic}[1]
\STATE \emph{Parameters} \\
 a. user selection probability $q \in (0, 1]$ \\
 b. per-user example cap $\hat{w} \in \mathbf{R}^+$ \\
 c. noise scale $z \in \mathbf{R}^+$ \\
 d. UserUpdate (for \texttt{FedAvg} or \texttt{FedSGD}) \\
 e. ClipFn (FlatClip or PerLayerClip) \\
 f. Parameter $W_{min}$ \\
\STATE \emph{Procedure:}
\STATE Initialize model $\theta^0$
\STATE $w_k =\min(\frac{n_k}{\hat{w}}, 1)$ for all users $k$
\FOR{each round $t = 0, 1, 2, \dots$}
\FOR{each zone $i = 1, 2, \dots, s$}
\STATE $\mathcal{C}_i^t \leftarrow$ (sample users with probability $q$)
\STATE $W = \sum_{k \in \mathcal{C}_i^t} w_k$ 
\FOR{each user $k \in \mathcal{C}_i^t$ \textbf{in parallel}}
\STATE $\Delta^{t+1}_k \leftarrow \text{UserUpdate}(k, \theta^t, \text{ClipFn})$ 
\ENDFOR
\STATE $\Delta(i)^{t+1} = 
 \begin{cases} 
 \frac{\sum_{k\in \mathcal{C}_i^t} w_k \Delta_k^{t+1}}{q \cdot W} 
 & \text{for FlatClip} \\[4pt]
 \frac{\sum_{k \in \mathcal{C}_i^t} w_k \Delta_k^{t+1}}{\max(q \cdot W_{min},\sum_{k\in \mathcal{C}_i^t} w_k)}
 & \text{for PerLayerClip}
 \end{cases}$
\STATE $\mathcal{S}_i \leftarrow $ (bound on $||{\Delta^{t+1}_k}||_2$ for ClipFn)
\STATE $\sigma_i \leftarrow \left\{\text{$\frac{z \mathcal{S}_i}{q \cdot W}$ for FlatClip or $\frac{2z\mathcal{S}_i}{q \cdot W_{min}}$ for PerLayerClip} \right\}$
\STATE $\theta^{t+1} \leftarrow \theta^t + \frac{1}{s} \sum_{i \in [s]}\left(\Delta(i)^{t+1} + \mathcal{N}(0, I\sigma_i^2)\right)$
\ENDFOR
\ENDFOR
\end{algorithmic}
\caption{Modified FL mechanism with zonal privacy. Detailed information about parameters and sub-procedures is provided in~\cite{mcmahan2017learning}.}
\label{alg1}
\end{algorithm}

\vspace{1mm}
\noindent{\bf Algorithm for Hierarchical FL with DP:} We summarize our approach in Algorithm~\ref{alg1}, which is heavily based on the approach taken by McMahan \etal~\cite{mcmahan2017learning}, with some key modifications to incorporate hierarchical FL for $s$ zones. From line 6, observe that gradient aggregation first occurs at a {\em zonal level}. This in turn leads to re-calibration of various parameters (lines 7-14) required to provide the DP guarantee\footnote{For more details on the various forms of clipping that can be employed, refer the original work~\cite{mcmahan2017learning}.}. Finally, the {\em noised zonal gradients} are averaged and aggregated in line 15. Note that in scenarios where {\em uniform weighting} is applied (\ie the contribution of each client is the same), $w_i=\frac{1}{k}$. Additionally, the variance of the noise to be added ($\sigma_i$ in line 14) is calculated as a function of the online clients {\em per zone} (as noted by he denominator term: $q \cdot W$ or $q \cdot W_{min}$), and not the total fraction of clients across all zones, as $W$ is re-calibrated based on the selected clients per round per zone. 

The proposed approach calculates the sensitivity based on the clipping bound $\mathcal{S}_i$ that is calculated across all clients across all zones; obtaining this information in practice will require additional communication between the super-nodes. Thus, this is approximated through the existence of a global clipping bound $C$ such that $C \geq \max_{i \in [s]} \mathcal{S}_i$.

\vspace{1mm}
\noindent{\bf Desiderata:} HDP should have the following properties: 
\begin{enumerate}
\itemsep0em
\item {\bf Trust assumptions:} Obtaining HDP requires selecting super-nodes, and how these are selected determines the trust assumptions needed. Ideally, {\em we should not add any unreasonable trust assumptions} (refer \S~\ref{sec:supernode_selection}).
\item {\bf Algorithmic correctness:} In the absence of noise addition, the models learnt using HFL must be the {\em same as in the status quo} (refer \S~\ref{sec:correctness}). 
\item {\bf Utility:} HDP should not significantly degrade the privacy provided by LDP, and not degrade the utility benefits provided by CDP. In particular, HDP should provide {\em advantageous privacy vs. utility trade-offs} (refer \S~\ref{sec:privacy}).
\end{enumerate}

\subsection{Threat Model}
\label{sec:threat}

We assume a minority of $\zeta$ ($1 \leq \zeta < k$) online clients are adversarial. The adversaries in our setup are honest-but-curious, \ie they can not deviate from protocol specifications. This is the commonly followed threat model in most prior works~\cite{melis2019exploiting,geiping2020inverting,zhu2020deep,wang2019beyond,yin2021see}. 
Their sole goal is to infer information about the training data of a target client (or group of clients). The only knowledge these adversaries are privy to are the gradient updates shared during training. In particular, the adversaries we consider are able to view:
\begin{enumerate}
\itemsep0em
\item any (gradient) updates {\em they} generate.
\item the joint update shared from the central aggregator.
\item the update generated at the {\em intermediary level} (\ie at the super-node), iff the adversary is located at said level. 
\end{enumerate}
Recent work~\cite{boenisch2021curious,pasquini2021eluding} formalize active adversaries against the FL ecosystem whose aim is to subvert different parts of the protocol. 
In both cases, effective noise addition can minimize attacker efficacy.  

\vspace{2mm}
\noindent{\bf Adversarial Goals.} We now describe two attacks that we consider in our work. These attacks enable the adversary to: (a) identify if a particular data-point (or a particular attribute within a data-point) was used during training\footnote{A dataset comprises of many data-points, each of which comprises of numerous attributes.}, or (b) try to reconstruct the data used for training by another client by observing the gradients shared.
These adversaries can utilize: (a) {data inference}~\cite{melis2019exploiting} where an adversarial client can calculate the update added to the (central) model by calculating the difference between its model state at the previous iteration and the current iteration (received from the central aggregator), and using knowledge of its own update to identify the aggregated update of all other clients. With this knowledge, this client can infer if a particular data-point (or attribute) was used or not (by one of the other clients). This approach can further be extended into the adversary attempting to obtain the value of a particular {\em attribute} used. These adversaries can also use (b) {data reconstruction}~\cite{geiping2020inverting,yin2021see,zhu2020deep}, where adversaries use a modified optimization problem which allows them to reconstruct an input (used to obtain a particular gradient).While theoretically sound for a batch size of 1, empirical results demonstrate unsuccessful reconstruction for larger batch sizes~\cite{geiping2020inverting,yin2021see}.

Note that the reconstruction attack can capture all effects of the inference attack; once the adversary has access to the exact data used for training, it can also infer if the data-point possesses specific attributes (or not). In our evaluation in \S~\ref{sec:eval}, we measure privacy through the privacy expenditure ($\varepsilon$) calculation, discuss under what circumstances data reconstruction may succeed, and evaluate hierarchical FL (with DP) under the stronger reconstruction attack. While such an attack is computationally more expensive than the inference attacks proposed in literature, they are more realistic; inference attacks assume that the adversary knows what it is looking for (\ie makes strong assumptions about adversarial knowledge), but this is often not the case.

\section{Designing Hierarchies}
\label{hierarchies}

As noted in \S~\ref{sec:overview}, our primary objective is to design hierarchies to understand if they induce advantageous privacy vs. utility trade-offs. However, the first problem encountered in designing hierarchies is {\em selecting the super-nodes}. This is an important problem as the super-node is responsible for adding the noise on behalf of all the federated clients it obtains the updates from. Since we assume an honest-but-curious model, all participants (including the selected super-nodes) do not deviate from the protocol. However, if clients are to share sensitive information (such as the updates produced as part of local learning), then they must trust the super-nodes in order to do so. Below, we describe {\em why} such trust may exist, and what these clients can do to amplify their trust (and minimize the exposure of their sensitive information).

\subsection{Choosing Super-Nodes}
\label{sec:supernode_selection}

\noindent{\bf 1. Exploiting Inherent Hierarchies:} Hierarchies exist in communication networks in the status quo~\cite{petrek2001large}; this information can be used to select super-nodes. For example, hierarchies are introduced by the offset of edge computing, where the edge-based base-stations~\cite{messous2017computation,liu2016paradrop} serve as an intermediary between the federated clients and the central aggregator. Traditionally, the manufacturer of various hardware components placed at the different conceptual levels are different~\cite{gueta2019sbft}. This allows us to assume that the federated clients, super-nodes, and central aggregator belong to different trust regions. However, protocols are run in software which are often proprietary to the party learning the model, which in this case is the actor at level 2. In such scenarios, it is unclear if the super-node and the aggregator lie in different trust regions. One could envision that the protocol is shared with a different party, and its execution is verified using techniques from cryptography~\cite{setty2012making}; this may alleviate some of the concerns associated with trust. 

\vspace{2mm}
\noindent{\bf 2. Elections:} Another approach is to {\em elect} the super-node, \ie the super-node is one of the federated clients. This too, ensures that the super-node is from a different trust region in comparison to the central aggregator.  First, federated clients are grouped into {\em zones}. For example, the grouping can be based on (a) geography, (b) compute capabilities, or (c) some form of structured or unstructured peer-to-peer overlay organization~\cite{lua2005survey-p2poverlays}. Then, a distributed election protocol is run to select a super-node for the particular zone. Methods for this problem can be drawn from past literature that studied it under the context of super-peer selection in P2P networks using different assumptions, metrics, etc.~\cite{lo2005super-peer-selection1, mahdy2007super-peer-selection2, li2019super-peer-selection3}. In our setting, such a process assumes that (a) the federated clients are aware of others who are active in that particular zone for the particular round, and (b) can communicate the outcome of the election between each other. While such assumptions introduce additional overheads in terms of communication, they are not unrealistic, as we will see soon. One key requirement in our setting is that the choice/election of the super-nodes is randomized \ie a particular client in a zone has bounded probability of being elected as the super-node, and each of the clients is equally likely to be elected. The elected super-node is ``in power'' for the duration it is online (and selected, across multiple learning rounds). Once the super-node is down, then a new one is elected to replace it. However, recall that a small fraction $\zeta$ of online clients (who are potential super-nodes) are malicious, and can subvert the election protocol through collusion. To this end, different election protocols and assumptions provide different guarantees. For example, to obtain conventional fault tolerance at zone $i$, a majority of the $k_i$ clients must be honest~\cite{castro2002practical} (\ie $\zeta_i < \frac{k_i-1}{2} \leq \zeta$).
Similarly, to obtain byzantine fault tolerance, $\zeta_i \leq \frac{k_i-1}{3} \leq \zeta$~\cite{castro2002practical}.

\vspace{2mm}
Formally, let us assume that there are a total of $s$ zones and each zone $i \in [s]$ has $n_i \leq n$ federated clients, such that $\sum_i n_i = n$.
All {\em online} clients in that particular round elect the super-node, \ie one of the $k_i \leq n_i$ online clients are elected as the super-node.
Note, however, that the elected super-node has complete purview to the gradients from {\em individual clients}, and these gradients may not be masked through the addition of noise needed for DP. To this end, we advocate for the usage of secure aggregation~\cite{bonawitz2017practical,bonawitz2016practical} protocols (within zones) to ensure that the super-node gets an aggregate view of the gradients from individual clients.
While recent work suggests that an actively malicious super-node (or central aggregator) can violate privacy guarantees provided by secure aggregation~\cite{boenisch2021curious, pasquini2021eluding}, this is out of the scope of our threat model\footnote{As we will explain in \S~\ref{subsec:fl_attacks}, such attacks can also be thwarted if the clients utilize LDP protocols.}
The work of Bonawitz \etal~\cite{bonawitz2016practical} suggests that (a) federated clients are aware of others who are online at the particular round to participate in secure aggregation protocols (similar to what is needed for distributed election protocols), and (b) the communication cost associated with secure aggregation is quadratic in the number of participants; both of these requirements can also be used to enable election protocols (as discussed earlier). In our setting, secure aggregation needs to be applied in all zones, and prior work suggests that secure aggregation protocols are practical (\ie introduces a tolerable time delay) for only a small number of clients~\cite{bonawitz2017practical}. This suggests that care must be taken in ensuring that the number of (online) clients per zone is also small. How this is achieved can be determined by future research.

\vspace{2mm}
\noindent{\bf Desiderata for Super-nodes:} We have described procedures to select super-nodes from the federated clients. Now, we describe desirable properties that the super-node should possess:
\begin{enumerate}
\itemsep0em
\item {\bf Compute capable:} The super-node should be capable of performing aggregation operations using the gradients it receives from the federated clients.
\item {\bf Synchronization:} Federated clients may be online at different times, and their communication with the super-node can be bottlenecked due to various network-related issues~\cite{awerbuch1985complexity}. The super-node should be capable of handling such distributed synchronization issues~\cite{arvind1994probabilistic,silberschatz1979communication}.
\item {\bf Persistence:} Elected super-nodes are privy to sensitive client information for the entire process. While this may result in a malicious client getting access to user data, the probability of this event is bounded, and the ill-effects can be minimized if the clients utilize LDP and secure aggregation to mask the shared updates. A naive alternative would be to elect a new super-node for every round. However, this would ensure that all the elected super-nodes are privy to sensitive user information (aggregated and noised, or otherwise). 
\end{enumerate}

\subsection{Algorithmic Correctness}
\label{sec:correctness}

In \S~\ref{sec:overview}, we discuss how intermediary super-nodes can be used to add noise on behalf of all online nodes in its corresponding zone.  Thus far, we have discussed how these super-nodes are selected. However, despite the presence of the intermediary level, algorithmic correctness must be preserved, \ie the aggregator has to ensure that the final value that it obtains (and propagates) is the same as that obtained in the baseline (\eg \texttt{FedAvg}~\cite{mcmahan2017communication}) scenario. 

To this end, we generalize the construction described earlier as follows: there are $n$ federated clients, and a total of $s$ zones (each with its own super-node). Let us assume that each zone has an equal number of clients $m$ such that $(m+1) \cdot s = n$. While this simplifying assumption enables easier calculation, this can be relaxed in practice. The federated clients within each zone clip their gradients (as is commonly done~\cite{mcmahan2017learning}), and set the clipping norm to a global constant, \ie all participants (at any level of the hierarchy) are aware of the clipping threshold. Recall that nodes in the zone utilize secure aggregation to share an aggregated view of the gradients with the super-node. Thus, each of the $s$ super-nodes has received an update from the corresponding online clients in the zone. These $s$ super-nodes will forward these updates to the central aggregator. In the baseline case, this intermediary-level does not exist (and consequently, such intermediary-level forwarding does not exist); noise addition (should the approach utilize CDP) occurs at the central aggregator. 

To achieve correctness (\ie ensure that the protocol is consistent with the baseline), the super-node can (a) apply noise required for DP as is to the aggregated gradients from each zone (and averaging by the total number of online clients for that round will occur at the central aggregator), or (b) average the gradients first (by the number of online clients in the zone)\footnote{We assume that the secure aggregation protocol returns just the sum, and not the average. The protocol can easily be modified to return the average as well.} before adding noise. Observe that the sensitivity of case (b) will be lower than that of case (a), because of the division required for averaging. Consequently, we advocate for averaging at the super-node following which noise is added. 

\subsection{Privacy Benefits}
\label{sec:privacy}

Having established that averaging before adding noise at the super-nodes is the right strategy (\S~\ref{sec:correctness}), we now discuss the benefits of such noise addition on utility.
We begin by introducing preliminaries of the Gaussian mechanism.
Note that the formalism below is borrowed from the work of Roth and Dwork~\cite{dwork2014algorithmic}.

\vspace{2mm}
\noindent{\bf 1. $\ell_2$-sensitivity:}
The $\ell_2$-sensitivity of $g:
\mathbf{N}^{|X|} \rightarrow
\mathbf{R}^k$ is:
$$\Delta_2 g = \max_{x,y \in \mathbf{N}^{|X|}} ||g(x) - g(y)||_2$$
where $||x-y||_1=1$, $||.||_p$ denotes the p-norm, and $X$ denotes some universe (such as the universe of all databases).

\vspace{2mm}
\noindent{\bf 2. Gaussian Mechanism (GM):} Let the privacy budget $\varepsilon \in (0,1)$. For some constant $c^2 > 2\log(\frac{1.25}{\delta})$, the GM with parameter $\sigma \geq \frac{c\Delta_2 g}{\varepsilon}$ is $(\varepsilon, \delta)$ DP. 

\vspace{2mm}
In the following lemma, we will describe how there exists a connection between LDP and CDP, if the GM is used to provide DP properties.
Note that this lemma is discussed in the absence of an intermediary level, for simplicity in explanation. 
Extending this to include the presence of super-nodes (or an intermediary level) is trivial. 

\begin{tcolorbox}
\noindent{\bf Lemma:} If each of $k$ (online) federated clients obtains $(\varepsilon_{LDP}, \delta)$ LDP using the GM, the central aggregator obtains $(\varepsilon_{CDP},\delta)$ DP, where $\varepsilon_{CDP} = \frac{\varepsilon_{LDP}}{\sqrt{k}}$. 
\end{tcolorbox}

\vspace{2mm}
\noindent{\em Proof:} Assume the existence of a global sensitivity bound $C$. Assume the existence of $k$ online clients, each of which utilize the GM to independently add noise to the gradients (denoted $g_j$ for $j \in [k]$) being computed to achieve $(\varepsilon_{LDP}, \delta)$ LDP guarantees. Thus, each client utilizes constants $c_j$ for $j \in [k]$ which satisfy the above condition (in the definition of the GM), and samples noise from distributions with standard deviation $\sigma_j$ for $j \in [k]$, \ie each client returns the following value: 
$$g_j + \eta_j, \forall j \in [k]$$ 
where $\eta_j \sim \mathcal{N}(0, \sigma_j^2), \forall j \in [k]$.

The central aggregator computes $\hat{g} = \frac{1}{k}\sum_{j=1}^k (g_j + \eta_j)$ (\eg as required by \texttt{FedAvg}~\cite{konevcny2016federated}). Note that $g_j$ is the gradient calculated by client $j$ multiplied by the size of client $j$'s dataset, and then divided by the total dataset size of all online clients.

We know that the sum of two Gaussian variables is Gaussian.
Extending this, we know that $\sum_{j=1}^k \eta_j = \eta$ since 

$$\sum_{j=1}^k \mathcal{N}(0, \sigma_j^2) = \mathcal{N}(0, \sum_{j=1}^k \sigma_j^2) = \mathcal{N}(0, \hat{\sigma}^2)$$
(where $\eta \sim \mathcal{N}(0, \hat{\sigma}^2)$).
We also know that 

$$\hat{\sigma}^2 = \sum_{j=1}^k \sigma_j^2 \geq \frac{\Delta_2f^2}{\varepsilon^2}\sum_{j=1}^k c_j^2$$

Setting $c_j=C$ and $\sigma_j = \sigma$ for all $j \in [k]$, we can observe that the central aggregator is $(\frac{\varepsilon_{LDP}}{\sqrt{k}},\delta)$ DP, if each client is $(\varepsilon_{LDP}, \delta)$ LDP.

\vspace{2mm}
\noindent{\textbf{Privacy Amplification:}} We can observe that by utilizing the GM for LDP as described above, the central aggregator is able to obtain a more {\em private} aggregation of the individual gradients than if it was to use the same parameters ($C$, $\sigma$). Such a result is commonly known as {\em privacy amplification} in DP literature, and in this case is a natural consequence of using the GM to provide LDP guarantees. Amplification is a technique where one amplifies a weak secret into a strong one, \ie the provided DP guarantee is stronger, with a comparable lesser amount of noise added. Amplification is highly beneficial for utility (since lesser noise is added, but strong privacy is obtained). Prior works observe such amplification effects through subsampling~\cite{balle2018privacy} (as used by the accountant to calculate tight bounds for privacy expenditure of algorithms like DP-SGD), and more recently through approaches such as random check-ins~\cite{balle2020privacy}.

\vspace{2mm}
\noindent{\bf Implications of Amplification:} The aforementioned lemma can be generalized to multiple levels of a hierarchy.
If level 0 utilizes the GM to obtain $(\varepsilon, \delta)$ LDP, then the intermediary level \ie level 1 (which aggregates the responses from $k_1$ level 0 clients) obtains $(\frac{\varepsilon}{\sqrt{k_1}},\delta)$ DP. Level 2 which aggregates responses from $k_2$ level 1 clients obtains $(\frac{\varepsilon}{\sqrt{k_1 \cdot k_2}},\delta)$ DP, all the way upto level $l$ (which aggregates the responses from $k_{l}$ clients from level $l-1$), which obtains $(\frac{\varepsilon}{\sqrt{\prod_{j=1}^l k_l}},\delta)$ DP. Observe that the privacy (a) is dependent on the lowest level where noise is added (which in this case, is level 0), and (b) becomes better as the number of levels increases (assuming $k_j > 1$ for all $j \in [l]$).
In \S~\ref{sec:eval}, we study the implications of such a mechanism on the utility of the learned classifier.

\vspace{2mm}
\noindent{\textbf{Amplification from Shuffling~\cite{bittau2017prochlo,erlingsson2019amplification}:}} In the status quo (\ie no hierarchies), the central aggregator is able to directly see which client shares a gradient (noisy or otherwise) with it. Shuffling is a protocol where this view is broken; the central aggregator no longer knows which client shares the gradient. Shuffling is often achieved through a shuffler, which can be instantiated by a mixnet, for example. Feldman \etal~\cite{feldman2022hiding} analyze the shuffling protocol, and quantify the privacy amplification it induces. Recall that the clients utilize a secure aggregation protocol (which utilizes some notion of shuffling) to share an aggregated gradient with the super-node (ergo, ensuring that the super-node does not have direct view of the gradients of individual clients). If the clients were to utilize a LDP protocol (for reasons we will explain in \S~\ref{sec:generalization}), then amplification by shuffling would exist (and averaging post aggregation is a post-processing step). For $k$ clients that utilize a LDP protocol before shuffling, the authors argue that this induces another $\frac{1}{\sqrt{k}}$ factor, resulting in $\varepsilon_{CDP} = \frac{\varepsilon_{LDP}}{k}$ (a combination of both shuffling-induced amplification, and amplification induced due to the sum of GMs). 

\vspace{2mm}
\noindent{\bf Note:} Feldman \etal~\cite{feldman2022hiding} prove their amplification result for a general DP mechanism. However, the amplification result we show in the lemma is a consequence of using the GM for DP. If the participants decide to use a different mechanism, then the lemma does not hold.

\vspace{2mm}
\noindent{\bf Consequence on Hierarchical FL:} Observe that the amplification result allows for stronger privacy (measured at the central aggregator) when noise is added at a lower level of the hierarchy (be it clients, super-nodes, or both). Note that the stage where the noise is added dictates the denominator term in the privacy budget (assuming amplification induced only due to the sum of GMs): $\sqrt{k}$ if noise is added at each of the clients, and $\sqrt{s}$ if noise is added at each of the super-nodes. Since $s < k$, the privacy budget when the noise is added at the super-nodes is more than when added at the client-level. Intuitively, this suggests that a classifier learnt using HDP is going to be more utilitarian than one learnt with LDP (as the amount of noise added is lower). In \S~\ref{sec:eval}, we empirically validate this claim.

\section{Generalization}
\label{sec:generalization}

\begin{table}[H]
\small
\centering
\begin{tabular}{c | c c c c}
\hline
& \textbf{Aggregator} & \textbf{Super-node} & \textbf{Client} & \textbf{Privacy} \\
\hline
{\bf C1} & \xmark & \xmark & \cmark & $\frac{\varepsilon}{\sqrt{k}}$\\
{\bf C2} & \xmark & \cmark & \xmark & $\frac{\varepsilon}{\sqrt{s}}$\\
{\bf C3} & \xmark & \cmark & \cmark & $\frac{\varepsilon}{\sqrt{\beta \cdot s \cdot m}} + \frac{\varepsilon}{\sqrt{(1 - \beta) \cdot s}}$\\
{\bf C4} & \cmark & \xmark & \xmark &${\varepsilon}$\\
{\bf C5} & \cmark & \xmark & \cmark & $\frac{\varepsilon}{\sqrt{\alpha.k}} + \varepsilon$\\
{\bf C6} & \cmark & \cmark & \xmark & $\frac{\varepsilon}{\sqrt{\beta.s}} + \varepsilon$\\
{\bf C7} & \cmark & \cmark & \cmark & $\frac{\varepsilon}{\sqrt{\beta \cdot s \cdot m}} + \frac{\varepsilon}{\sqrt{\alpha \cdot s}} + \varepsilon$\\
\hline
\end{tabular}
\caption{Different configurations that are possible with our proposed hierarchical scheme. We will explain when these cases are realizable in \S~\ref{sec:generalization}.}
\label{tab:configurations}
\end{table}

In this section, we provide a generalization of the technique discussed thus far, with the corresponding privacy budgets. Assume a total of $s$ zones (as defined earlier) such that each zone $i$ has $n_i$ clients (and $\sum_{i \in [s]} n_i = n$). To make discussion easier, let us assume $n_i = m$ (for all $i$), and the total number of online clients per round is $k$.

\vspace{2mm}
\noindent{\bf C1} corresponds to the setting of pure LDP when {\em all clients} add the required noise to obtain ${\varepsilon}$ LDP. From the central aggregator's perspective, the overall privacy is $\frac{\varepsilon}{\sqrt{k}}$. This is visualized in Figure~\ref{fig:architecture-overview}, towards the far right (in the absence of secure aggregation).

\vspace{2mm}
\noindent{\bf C2} corresponds to the setting of pure HDP when {\em all super-nodes} add the required noise to obtain ${\varepsilon}$ HDP. From the central aggregator's perspective, the overall privacy is $\frac{\varepsilon}{\sqrt{s}}$. This is visualized in Figure~\ref{fig:architecture-overview}, in both zones 2 and 3 (in the absence of secure aggregation).

\vspace{2mm}
\noindent{\bf C3} is realizable when a fraction of all online clients choose to add the required noise for LDP, but the remainder do not. To ensure that the remaining clients receive privacy, the super-node corresponding to these clients needs to add noise to ensure DP. To simplify the math, let us assume that clients corresponding to $\beta \cdot s$ super-nodes choose to utilize LDP and the remaining $(1-\beta)\cdot s$ super-nodes provide HDP. If each zone has $m$ clients, then at each of the $\beta.s$ super-nodes, we observe $\frac{\varepsilon}{\sqrt{m}}$ DP. From the central aggregator's perspective, the total privacy is 
$$\frac{\varepsilon}{\sqrt{\beta \cdot s \cdot m}} + \frac{\varepsilon}{\sqrt{(1 - \beta) \cdot s}}$$ where the second term in the sum is the contribution of the $(1-\beta).s$ super-nodes that provide $\varepsilon$ HDP.  

\vspace{2mm}
\noindent{\bf C4} corresponds to the setting of pure CDP when {\em only the central aggregator} adds noise to obtain ${\varepsilon}$ CDP. This is visualized in Figure~\ref{fig:architecture-overview}, towards the far left.

\vspace{2mm}
\noindent{\bf C5} corresponds to the setting where a fraction $\alpha$ of clients do not trust the central aggregator and consequently wish to achieve $\varepsilon$ LDP. To provide privacy to the remainder, the central aggregator also adds noise. Thus, the total privacy budget from the central aggregator's perspective is $$\frac{\varepsilon}{\sqrt{\alpha.k}} + \varepsilon$$ 
where the first term is due to the $\alpha$ fraction which achieves LDP.

\vspace{2mm}
\noindent{\bf C6} is similar to {\bf C5}, in that only a fraction $\beta$ of super-nodes add noise to achieve $\varepsilon$ HDP. To provide privacy to the remainder, the central aggregator also adds noise. Thus, the total privacy budget from the central aggregator's perspective is $$\frac{\varepsilon}{\sqrt{\beta.s}} + \varepsilon$$
where the first term is due to the $\beta$ fraction which achieves HDP.

\vspace{2mm}
\noindent{\bf C7} corresponds to the scenario where (a) clients in $\beta.s$ zones do not trust any entity and wish to achieve $\varepsilon$ LDP, (b) clients in $\alpha.s$ zones trust the super-nodes, and thus the super-nodes achieves $\varepsilon$ HDP, and (c) for the clients in the remaining $(1-\alpha-\beta)\cdot s$ zones, the central aggregator is responsible for providing privacy. Thus, from the central aggregator's perspective, the total privacy is $$\frac{\varepsilon}{\sqrt{\beta \cdot s \cdot m}} + \frac{\varepsilon}{\sqrt{\alpha \cdot s}} + \varepsilon$$
where the first term is due to (a), and the second term is due to (b).

\vspace{2mm}
\noindent{\bf Note:} All the calculations above are in the absence of the amplification due to secure aggregation. Should it be considered, the only change would be a removal of the square root over all terms.

\section{Implementation}
\label{sec:implementation}

We implement our proposed approach using \texttt{tensorflow}.
In particular, we use a combination of \texttt{tensorflow-federated v0.16.1} to provide the components required for FL, and \texttt{tensorflow-privacy} to provide the machinery required for private learning\footnote{\url{https://github.com/tensorflow/federated/tree/v0.16.1} and \url{https://github.com/tensorflow/privacy}}.
In particular, such a framework allows us to calibrate for the clipping norm $C$ and the noise mulitplier~\cite{abadi2016deep}; modifications to either will result in implications to both accuracy and privacy. To ensure correct accounting, we modify the accounting libraries in \texttt{tensorflow-privacy} to accurately reflect sampling probability and number of iterations (in this case, rounds). 

To evaluate the efficacy of our approach, we consider the datasets listed in Table~\ref{tab:dataset}; only the EMNIST dataset is modified to exhibit the non-i.i.d property that is commonly associated with FL. Note that the objective of our evaluation is to understand the advantageous utility vs. privacy trade-offs introduced by our scheme. To this end, how non-i.i.d the data is distributed between the clients is not an important compounding factor. The evaluation we will discuss can be considered as an average-case evaluation of the proposed hierarchical DP approach.

\begin{table}[H]
\small
    \centering
    \begin{tabular}{l*{4}c}
        \hline
        \textbf{Dataset} & \textbf{Size} & \textbf{\# Entries} & \textbf{\# Classes} & \textbf{$n$} \\
        \hline
        \text{EMNIST~\cite{cohen2017emnist}} & $28\times28\times1$ & 382705 & 10 & 3383\\
        \text{CIFAR-10~\cite{cifar10}} & $32\times32\times3$ & 60000 & 10 & 500 \\
        \text{CIFAR-100~\cite{cifar100}} & $32\times32\times3$ & 60000 & 100 & 500 \\
        \hline
    \end{tabular}
    \caption{Dataset characteristics.}
    \label{tab:dataset}
\end{table}

The datasets follow a standard 80:20 split (80\% of the data is used for training, and the remaining is used for validation). All our experiments were executed on a server with 2 NVIDIA Titan XP with 128 GB RAM and 48 CPU cores running Ubuntu 20.04.2. Due to computational constraints, and issues in \texttt{tensorflow-federated} related to GPU execution\footnote{\url{https://github.com/tensorflow/federated/issues/832}}, our experimental setup is conservative; we only perform a single run of each configuration and are unable to report error bars. However, we replicate the ecosystem chosen in prior work~\cite{mcmahan2017learning}.

For all experiments (unless explicitly specified otherwise), we choose a setup where $k=100$ users are randomly sampled per federated round (this amounts to different values of the sampling/user-selection probability for different datasets). \texttt{FedAvg}~\cite{mcmahan2017communication} is used as the algorithm of choice, and each client performs local training for 5 epochs. FL is performed for 200 rounds. All participating clients use SGD as the learning algorithm, with a learning rate of 0.02 (this includes clients that are super-nodes). The server's learning rate is set to 1. These parameters were chosen based on the prescribed guidelines from the authors of \texttt{tensorflow-federated} and from prior work~\cite{mcmahan2017communication,mcmahan2017learning}.

We re-ierate that the objective of our evaluation is centered around demonstrating the benefits of a hierarchical approach for learning in terms of the impact it has on the privacy vs. utility trade-off. To this end, we choose representative architectures from prior work for the datasets we consider~\cite{mcmahan2017communication,mcmahan2017learning}. In particular, we consider (a) a shallow 1 hidden layer DNN for EMNIST, and (b) 2 convolution layers followed by 2 fully connected layers for both CIFAR-10 and CIFAR-100. We will release all code used for our experiments on request.

\section{Evaluation}
\label{sec:eval}

Having described our evaluation setup in \S~\ref{sec:implementation}, we now focus on the results. Through our evaluation, we wish to answer the following questions:
\begin{enumerate}
\itemsep0em
\item Does the proposed hierarchical scheme provide advantageous privacy vs. utility trade-offs in comparison to the baseline scenarios of using (a) LDP and/or (b) CDP?\footnote{Recall that LDP provides more privacy, while CDP is known to provide a more utilitarian model~\cite{DBLP:journals/corr/abs-2102-05975}.} 
\item Does the proposed hierarchical scheme create a new attack surface for an adversary wishing to perform data (or membership) inference?
\end{enumerate}
Recall from \S~\ref{sec:privacy}, that such a distributed learning framework (with the GM for DP) leads to privacy amplification (\ie privacy when viewed through higher levels is different than what is viewed through at lower levels of the hierarchy). Thus, to ensure a fair comparison, we only report the privacy budget ($\varepsilon$) calculated at the central aggregator level (\ie $\varepsilon_{CDP}$) in our results. Our results suggest that:
\begin{enumerate}
\itemsep0em
\item {\em The theoretical intuition is corroborated in empirical measurements of utility}. As expected, the proposed approach provides an advantageous trade-off between utility and privacy, and outperforms the LDP baseline in scenarios where the number of super-nodes ($s$) is lesser than the number of online clients ($k$).
\item Hierarchical FL is also more resilient to data reconstruction attacks~\cite{zhu2020deep}. In the scenario where the super-node is adversarial, then it is able to perform perfect reconstruction (under some assumptions about the batch-size of data used for client-side learning, and/or in the absence of secure aggregation). However, if the client is adversarial, we observe that reconstruction efforts are less successful than in the scenario with central DP.
\end{enumerate}

\subsection{Privacy vs. Utility Trade-Off}

We first train without DP to get an estimate of the $2-$norm of the gradients. This helps us estimate the clipping norm ($C$) to be used during DP training. We also observe the training duration (\ie exact epoch number) at which the validation accuracy saturates. Once this is obtained, we configure the noise multiplier ($z$) so as to enable DP training and log the privacy expenditure ($\varepsilon$) at the end of training. Note that our training results, configurations, datasets and models are similar to those of prior work in this area~\cite{mcmahan2017learning}.

\begin{figure*}
\centering
\begin{subfigure}[b]{0.32\textwidth}
   \includegraphics[width=1\linewidth]{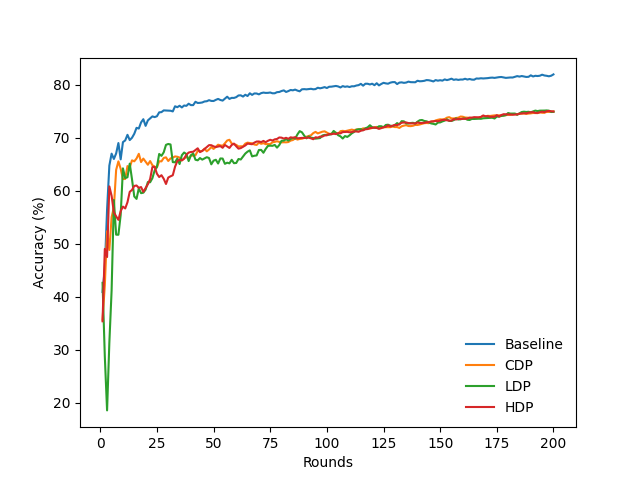}
   \caption{EMNIST}
   \label{fig:2a} 
\end{subfigure}
\begin{subfigure}[b]{0.32\textwidth}
   \includegraphics[width=1\linewidth]{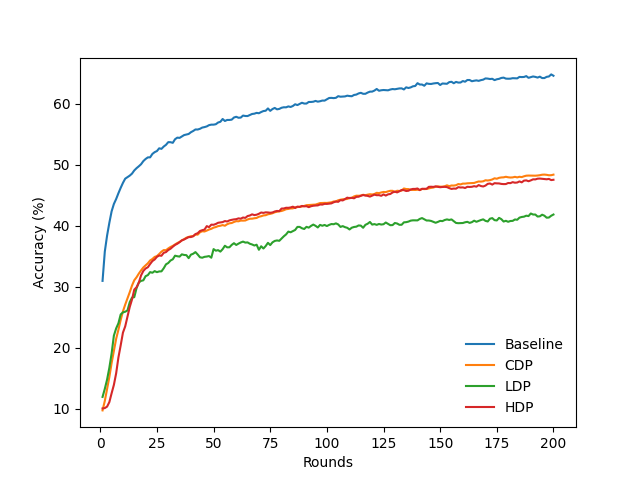}
   \caption{CIFAR-10}
   \label{fig:2b}
\end{subfigure}
\begin{subfigure}[b]{0.32\textwidth}
   \includegraphics[width=1\linewidth]{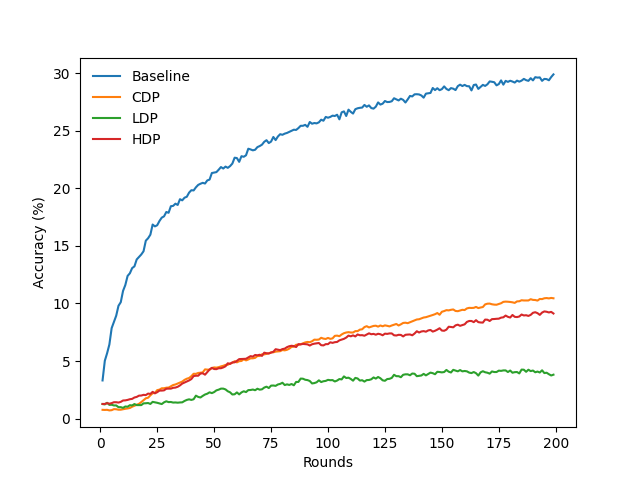}
   \caption{CIFAR-100}
   \label{fig:2c}
\end{subfigure}
\caption[]{We plot the validation accuracy of training with differential privacy for 3 scenarios: (i) central DP or CDP (in orange), (ii) local DP or LDP (in green), and (iii) our proposed hierarchical DP or HDP (in red), in comparison to training without privacy (in blue).
This is done across the three datasets: (a) EMNIST, (b) CIFAR-10 and (c) CIFAR-100.
Observe that HDP achieves better utility than the LDP scenario, and is often close to the CDP case across datasets.}
\label{fig:2}
\end{figure*}

\vspace{1mm}
\noindent{\bf Privacy vs. Utility:} Observe that different strategies of applying DP noise (\ie central vs. local vs. hierarchical) result in differing values of privacy expenditure. To ensure a fair comparison of privacy vs. utility, we convert the privacy expenditure in all settings to that of the central DP privacy expenditure (using the formulation presented in \S~\ref{sec:privacy})\footnote{Note that the $\varepsilon$ values we report are the ones obtained without the amplification due to shuffling; secure aggregation is known to be communication inefficient~\cite{bonawitz2016practical,bonawitz2017practical} and we wish to provide insight on the privacy vs. utility trade-offs in its absence as an average case estimate.}. We then plot the validation accuracy as a function of training duration for all the datasets we consider. We also consider a scenario without any form of DP training enabled; this is our {\em baseline}. Note that for the hierarchical DP setting, we consider a scenario where there are $s=10$ super-nodes. The results are plotted in Figure~\ref{fig:2}.

Observe that learning simple tasks such as EMNIST (refer Figure~\ref{fig:2a}) with DP is achievable in all three settings; prior work has demonstrated that learning with privacy for EMNIST can be as performant as learning without privacy~\cite{mcmahan2017learning} \ie the accuracy degradation induced by LDP in comparison to CDP is minimal. However, as expected, HDP provides advantageous trade-offs in terms of privacy (\ie the privacy budget is lower than CDP for comparable accuracy). 

The results are more interesting for complex datasets such as CIFAR-10 and CIFAR-100. First, observe that for the particular configuration we choose (\ie $n=500$, $k=100$), baseline accuracy is $\sim63\%$ and $\sim28\%$ for the two CIFAR versions, respectively. Note that these values are comparable to those achieved by McMahan \etal (refer Fig. 4 in~\cite{mcmahan2017learning}). Training with CDP degrades this accuracy further. However, as expected, HDP is able to provide advantageous trade-offs in terms of privacy and accuracy. Note that CIFAR-100 is naively at least $10\times$ more complex a learning task than CIFAR-10, and yet (a) HDP achieves similar utility to CDP in this setting, and (b) much higher utility than LDP.

Finally, Table~\ref{tab:privacy} contains corresponding values of the privacy budget ($\varepsilon$) achieved at the end of training. Observe that the privacy budget calculated at the end of HDP training is in-between that of the CDP and LDP case. In fact, with nearly a 67\% decrease in privacy expenditure, HDP is able to achieve nearly the same accuracy as the CDP case {\em across all 3 datasets} we consider.

\begin{table}
\small
    \centering
    \begin{tabular}{l*{3}c}
        \hline
        \textbf{Dataset} & \textbf{LDP} & \textbf{HDP} & \textbf{CDP}\\
        \hline
        \text{EMNIST} & 0.30 & 0.96 & 3.06\\
        \text{CIFAR-10} & 2.48 & 7.48 & 24.80 \\
        \text{CIFAR-100} & 2.48 & 7.48 & 24.80 \\
        \hline
    \end{tabular}
    \caption{Privacy Expenditure across datasets and DP methods. Note that the values for CIFAR-10 and CIFAR-100 are the same as the values of $n$ and $k$ are the same in both settings, and these datasets have the same number of data-points and are trained for the same duration.}
    \label{tab:privacy}
\end{table}

\begin{figure*}
\centering
\begin{subfigure}[b]{0.32\textwidth}
   \includegraphics[width=1\linewidth]{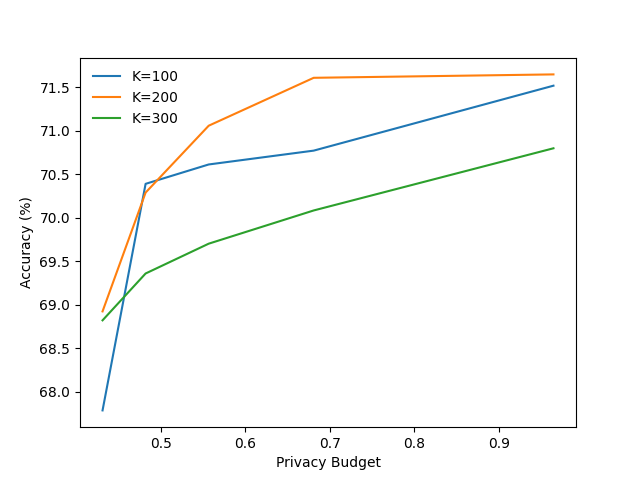}
   \caption{EMNIST}
   \label{fig:4a} 
\end{subfigure}
\begin{subfigure}[b]{0.32\textwidth}
   \includegraphics[width=1\linewidth]{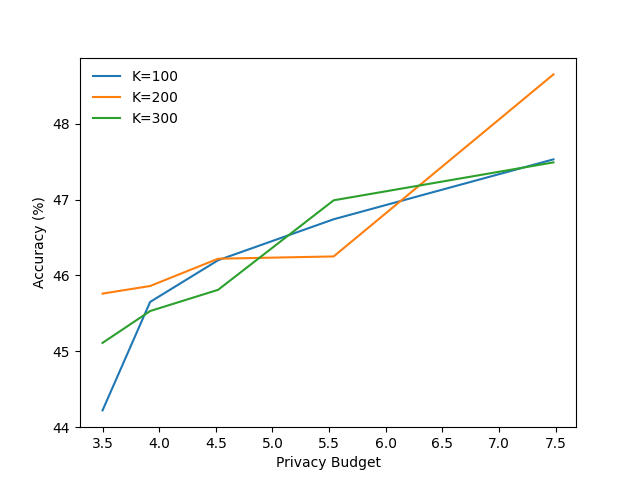}
   \caption{CIFAR-10}
   \label{fig:4b}
\end{subfigure}
\begin{subfigure}[b]{0.32\textwidth}
   \includegraphics[width=1\linewidth]{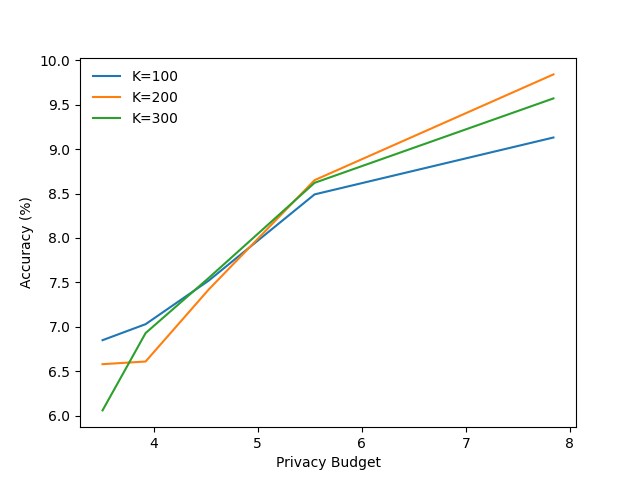}
   \caption{CIFAR-100}
   \label{fig:4c}
\end{subfigure}
\caption[]{We plot the validation accuracy as a function of privacy ($\varepsilon$) (obtained by varying the value of $s$) for 3 scenarios using HDP: (i) $k=100$ (in blue), (ii) $k=200$ (in orange), and (iii) $k=300$ (in green), across the three datasets: (a) EMNIST, (b) CIFAR-10, and (c) CIFAR-100. Observe that the utility obtained by HDP improves with increasing privacy budget $\varepsilon$.}
\label{fig:4}
\end{figure*}

\noindent{\bf The Influence of $s$:} From our analysis, we can see that increasing the number of super-nodes ($s$) makes the scheme more private (the privacy budget is inversely proportional to the value of the number of parties where the noise is being added; in the case of HDP, this is $s$). To better understand this hyperparameter, we consider an experimental setting where we vary the value of $k$ to 100, 200, and 300. Across these 3 settings, we vary the value of $s$ to one of $\{10,20,30,40,50\}$ across all datasets. All other hyperaprameters were kept the same as in the earlier experiment. We then measure the validation accuracy and vary the configurations of $C$ and $z$ to obtain the privacy expenditure. We plot the relationship between the fully trained model's validation accuracy and the privacy budget expended to achieve it in Figure~\ref{fig:4}. Across all datasets we observe a common trend: the validation accuracy calculated increases as the privacy expenditure does. For the datasets we consider, increasing the value of $k$ does not increase the validation accuracy substantially. This is {\em not} indicative of a more general trend; one would assume that increasing the number of participants would result in a more accurate model.

\begin{tcolorbox}
\noindent{\bf Take-away:} The proposed hierarchical approach provides advantageous privacy vs. utility trade-offs in comparison to both CDP and LDP-based approaches. Additionally, increasing the value of $s$ provides better privacy, which in-turn leads to lower utility. 
\end{tcolorbox}

\subsection{Attacks on Federated Learning}
\label{subsec:fl_attacks}

We wish to understand if the proposed hierarchical scheme introduces a new attack surface: through the introduction of adversarial entities at the super-node level. In this subsection, we will discuss (a) the capability of attackers in the status quo, (b) if the success of attacks in the status quo increases in the new hierarchical scheme, and (c) if any new attacks are possible due to the introduction of hierarchies.

\begin{table}[H]
\small
    \centering
    \begin{tabular}{l*{4}c}
        \hline
        \textbf{Dataset} & \textbf{LDP} & \textbf{HDP} & \textbf{CDP} & \textbf{No DP}\\
        \hline
        \text{EMNIST} & 0.571 & 0.599 & 0.62 & 0.68\\
        \text{CIFAR-10} & 0.423 & 0.452 & 0.494 & 0.596\\
        \text{CIFAR-100} & 0.447 & 0.46 & 0.474 & 0.564\\
        \hline
    \end{tabular}
    \caption{Efficacy of data reconstruction: Observe that reconstruction is least effective when LDP is used for DP training; HDP provides the next best resilience.}
    \label{tab:lpips}
\end{table}

\noindent{\bf Level 0 Adversaries:} Here, we consider scenarios where the adversaries are at level 0, and aim to perform data reconstruction (of a particular local client). Recall that in FL, the central aggregator adds the aggregated (client or super-node calculated) gradients to its own weights before propagating a new updates for the next round. Based on the generalization proposed in \S~\ref{sec:generalization}, one can observe that the noise addition required to provide DP can occur at (one or all of the) three levels: (a) the clients themselves add DP noise (\ie a pure LDP scheme), (b) the central aggregator adds noise (\ie a pure CDP scheme), and (c) the super-nodes add noise (\ie a HDP scheme). Regardless of which level adds the noise, we assume that noise addition is performed and a malicious client (at level 0) receives the {\em noisy update} from the central aggregator, and wishes to use this information to enable data reconstruction. To do so, the client is able to subtract its contribution from the aggregated weight update shared, and run reconstruction attacks using the remaining information (the strategy proposed by Melis \etal~\cite{melis2019exploiting}). To perform reconstruction, we implement the attack proposed by Zhu \etal~\cite{zhu2020deep}, using the source code presented by the authors for the datasets and models considered in the earlier section. We measure the reconstruction capabilities using the Learned Perceptual Image Patch Similarity (LPIPS) metric~\cite{zhang2018unreasonable} (larger values are more realistic). As prior work notes, such reconstruction attacks are largely inefficient for large batch sizes (used for local learning) and large values of $k$. 

To simplify the setup, and highlight the merit of our scheme, we consider a simple setup of $s=2$ and $k=4$, and where one of these clients is adversarial and wishes to learn the data of the other, when the batch size used is $1$. The results are presented in Table~\ref{tab:lpips}. Observe that the HDP scheme provides better resilience than CDP, but is worse than the LDP scheme. This is explained by the privacy guarantees provided by HDP (which, again, lies between LDP and CDP).

\vspace{2mm}
\noindent{\bf Level 1 adversaries:} Privacy attacks at the super-node level can be caused by the adversary having complete purview to the gradients from each client in that zone. While we advocate for the usage of secure aggregation to alleviate this issue, we discuss the outcomes if such a protocol can not be deployed. In such situations, as denoted by Figure~\ref{fig:architecture-overview}, the super-node can have direct access to gradients from individual clients and can perform data reconstruction attacks (to determine membership). We describe what may happen in such scenarios:
\begin{itemize}
\itemsep0em
\item In the pure LDP setting (\ie {\bf case 1} from Table~\ref{tab:configurations}), federated clients add noise to the gradients shared upstream. Thus, the super-node adversary has a noisy view of each of the per-client gradient it receives. In such settings, attacks such as the one by Melis \etal~\cite{melis2019exploiting} will be rendered ineffective (as this attack requires exact gradient knowledge), and data reconstruction attacks become less effective, as these attacks rely on correct gradient information for reconstruction (refer Table~\ref{tab:lpips}). 
\item In scenarios where federated clients {\em do not} add noise (\ie {\bf case 2} from Table~\ref{tab:configurations}), and the super-node is required to perform noise addition (as in the of HDP), data reconstruction attacks are possible if the super-node is malicious as it has direct purview to the gradients from the local clients. However, if secure aggregation is utilized, the malicious super-node is provided an {\em aggregated} gradient; the efficacy of these attacks in such cases is reduced~\cite{geiping2020inverting,yin2021see} \ie reconstruction is not of high fidelity (refer to the {\em No DP} column in Table~\ref{tab:lpips}).
\end{itemize}

To make level 1 adversaries less effective, (a) either client-level noise needs to be added, or (b) an aggregated gradient (via secure aggregation) needs to be shared to the super-node. Local DP (\ie client-level noise addition) is a very computationally efficient procedure, and its impacts on utility are well understood (\ie it induces an unfavorable utility vs. privacy trade-off). Protocols like secure aggregation, on the other hand, do not harm the utility of the model (\ie the utility of the model in the absence of secure aggregation is the same with it). However, it greatly increases the communication cost associated with the protocol; each client needs to communicate with all other clients (for each aggregation, in the worst case), leading to communication complexity that is quadratic in the number of clients. We measured the time required to perform secure aggregation using the \texttt{tff.learning.secure\_aggregator} module available as part of \texttt{tensorflow-federated}, and for the models we described earlier. Per round, we needed between 2 and 30 seconds (the time increases as a function of the model size). These numbers were consistent with the numbers presented earlier~\cite{bonawitz2016practical,bonawitz2017practical}.

\vspace{2mm}
\noindent{\bf Level 3 adversaries:} If the central aggregator is malicious, the setting is the same as the level 2 adversary. Both LDP and HDP in conjunction with secure aggregation can reduce such an adversary's efficacy of performing data reconstruction. However, there is a strategic advantage of performing secure aggregation at the super-node level. In a simple setting, assume the existence of $s$ zones each with $m$ clients, and $s << k$ (where $k$ is the total number of online clients). It is clear to see how the communication cost associated with secure aggregation is lower in this setting (it is quadratic in $s$ and not $k$).

\begin{tcolorbox}
{\bf Take-away:} In addition to secure aggregation, (a) to defeat level 0 adversaries, either LDP or HDP suffices; (b) to defeat level 1 adversaries, LDP suffices; and (c) to defeat level 2 adversaries, LDP or HDP suffices. However, HDP provides better utility than LDP, and is preferred wherever possible. 
\end{tcolorbox}

\vspace{2mm}
\noindent{\bf Note:} We consider passive adversaries which aim to perform reconstruction, as done in prior work. More recent work~\cite{boenisch2021curious,pasquini2021eluding} considers actively malicious actors that can provide inconsistent information to different actors of the FL ecosystem to facilitate more efficient data reconstruction (without being influenced by the batch size). However, the efficacy of such attacks are reduced when secure aggregation is combined with noise addition (required to provide DP). 
\section{Discussion \& Open Questions}
\label{sec:discussion}

\noindent{\bf Influence of $k$:} The privacy accounting in federated learning stems from multiple factors, primary of which is the sampling probability; this determines the number of clients that are online in each federated round. Contrary to our expectation, our experimental results suggest that increasing the value of $k$ does not have a strong effect on the accuracy of the final model learnt. We conjecture that this maybe the case due to the i.i.d distribution of data associated with both the CIFAR-10 and CIFAR-100 datasets. While EMNIST is non-i.i.d, the learning task by itself is too simple to merit substantial utility difference across the three settings we consider. However, a small value for the sampling probability \ie a small value of $k$ greatly improves the privacy of the model learnt (\ie small values of $\varepsilon$). We leave performing a more in-depth analysis of the influence of $k$ and $\epsilon$ for future work.

\vspace{1mm}
\noindent{\bf Privacy Amplification:} New approaches for privacy amplification (such as randomized check-ins~\cite{balle2020privacy}) can easily be incorporated with our scheme\footnote{Depending on the level it is applied, we will obtain a different amplification factor}. Oftentimes, amplification is a byproduct of composing a process that provides randomization with the actual function that needs to be made differentially private. In our scheme, such a randomization effect is observed through secure aggregation (which enables shuffling). In the scenario where the $s=k$, there is no amplification provided as the central aggregator views individual client gradients. However, when $s<k$, observe that the central aggregator is only provided an aggregate view (of all gradients from a particular zone). Determining if other sources of amplification may exist is subject to future work.

\vspace{1mm}
\noindent{\bf Information Leakage:} If the number of clients per zone exceeds the number of federated rounds, then in expectation, each client will be elected a super-node only once. A more detailed understanding is needed to ascertain how much information is leaked by exposing a super-node {\em only once} to aggregate gradient information multiple times. 

\vspace{1mm}
\noindent{\bf Runtime:} In our work, we measure the privacy vs. utility trade-offs of the proposed hierarchical ecosystem, assuming that elections and secure aggregation are implemented using state-of-the-art approaches. Microbenchmarks providing run-times of the overall scheme, highlighting time taken for both elections and secure aggregation, as a function of both $n$ and $k$, will help understand the practicality of the scheme.

\vspace{1mm}
\noindent{\bf Practical Topologies \& Data Distribution:} In our current work, we prototype the proposal using simple topologies where each super-node is responsible for the same number of clients. One can theoretically show that varying the number of clients per super-node will not influence the privacy guarantees of the approach. However, this will have an impact on the utility of the final model learnt. 
\section{Conclusion}

In our work, we propose an approach for hierarchical federated learning through super-node election from federated clients. We also propose extensions for how it can be retrofitted to provide differential privacy guarantees. Our experiments suggest that the proposed approach provides more advantageous privacy vs. utility trade-offs compared to approaches in the status quo, while providing resilience to inference adversaries. In future work, we hope to analyze how the randomization involved in super-node election can be composed with the amplification provided by the Gaussian mechanism to provide stronger privacy guarantees.

\newpage
\bibliographystyle{unsrt}
\bibliography{biblio,new}

\begin{thebibliography}{10}

\bibitem{mcmahan2017communication}
Brendan McMahan, Eider Moore, Daniel Ramage, Seth Hampson, and Blaise~Aguera
  y~Arcas.
\newblock Communication-efficient learning of deep networks from decentralized
  data.
\newblock In {\em Artificial Intelligence and Statistics}, pages 1273--1282.
  PMLR, 2017.

\bibitem{melis2019exploiting}
Luca Melis, Congzheng Song, Emiliano De~Cristofaro, and Vitaly Shmatikov.
\newblock Exploiting unintended feature leakage in collaborative learning.
\newblock In {\em 2019 IEEE Symposium on Security and Privacy (SP)}, pages
  691--706. IEEE, 2019.

\bibitem{boenisch2021curious}
Franziska Boenisch, Adam Dziedzic, Roei Schuster, Ali~Shahin Shamsabadi, Ilia
  Shumailov, and Nicolas Papernot.
\newblock When the curious abandon honesty: Federated learning is not private.
\newblock {\em arXiv preprint arXiv:2112.02918}, 2021.

\bibitem{geiping2020inverting}
Jonas Geiping, Hartmut Bauermeister, Hannah Dr{\"o}ge, and Michael Moeller.
\newblock Inverting gradients--how easy is it to break privacy in federated
  learning?
\newblock {\em arXiv preprint arXiv:2003.14053}, 2020.

\bibitem{yin2021see}
Hongxu Yin, Arun Mallya, Arash Vahdat, Jose~M Alvarez, Jan Kautz, and Pavlo
  Molchanov.
\newblock See through gradients: Image batch recovery via gradinversion.
\newblock {\em arXiv preprint arXiv:2104.07586}, 2021.

\bibitem{zhu2019federated}
Wennan Zhu, Peter Kairouz, Haicheng Sun, Brendan McMahan, and Wei Li.
\newblock Federated heavy hitters discovery with differential privacy.
\newblock {\em arXiv preprint arXiv:1902.08534}, 2019.

\bibitem{pasquini2021eluding}
Dario Pasquini, Danilo Francati, and Giuseppe Ateniese.
\newblock Eluding secure aggregation in federated learning via model
  inconsistency.
\newblock {\em arXiv preprint arXiv:2111.07380}, 2021.

\bibitem{dwork2014algorithmic}
Cynthia Dwork, Aaron Roth, et~al.
\newblock The algorithmic foundations of differential privacy.
\newblock {\em Foundations and Trends in Theoretical Computer Science},
  9(3-4):211--407, 2014.

\bibitem{abadi2016deep}
Martin Abadi, Andy Chu, Ian Goodfellow, H~Brendan McMahan, Ilya Mironov, Kunal
  Talwar, and Li~Zhang.
\newblock Deep learning with differential privacy.
\newblock In {\em Proceedings of the 2016 ACM SIGSAC Conference on Computer and
  Communications Security}, pages 308--318, 2016.

\bibitem{chaudhuri2011differentially}
Kamalika Chaudhuri, Claire Monteleoni, and Anand~D Sarwate.
\newblock Differentially private empirical risk minimization.
\newblock {\em Journal of Machine Learning Research}, 12(Mar):1069--1109, 2011.

\bibitem{10.1145/3035918.3064047}
Xi~Wu, Fengan Li, Arun Kumar, Kamalika Chaudhuri, Somesh Jha, and Jeffrey
  Naughton.
\newblock Bolt-on differential privacy for scalable stochastic gradient
  descent-based analytics.
\newblock In {\em Proceedings of the 2017 ACM International Conference on
  Management of Data}, SIGMOD '17, page 1307–1322, New York, NY, USA, 2017.
  Association for Computing Machinery.

\bibitem{song2021systematic}
Liwei Song and Prateek Mittal.
\newblock Systematic evaluation of privacy risks of machine learning models.
\newblock In {\em 30th $\{$USENIX$\}$ Security Symposium ($\{$USENIX$\}$
  Security 21)}, 2021.

\bibitem{geyer2017differentially}
Robin~C Geyer, Tassilo Klein, and Moin Nabi.
\newblock Differentially private federated learning: A client level
  perspective.
\newblock {\em arXiv preprint arXiv:1712.07557}, 2017.

\bibitem{naseri2020toward}
Mohammad Naseri, Jamie Hayes, and Emiliano De~Cristofaro.
\newblock Toward robustness and privacy in federated learning: Experimenting
  with local and central differential privacy.
\newblock In {\em 29th Network and Distributed System Security Symposium
  (NDSS)}, 2022.

\bibitem{konevcny2016federated}
Jakub Kone{\v{c}}n{\`y}, H~Brendan McMahan, Felix~X Yu, Peter Richt{\'a}rik,
  Ananda~Theertha Suresh, and Dave Bacon.
\newblock Federated learning: Strategies for improving communication
  efficiency.
\newblock {\em arXiv preprint arXiv:1610.05492}, 2016.

\bibitem{kasiviswanathan2011can}
Shiva~Prasad Kasiviswanathan, Homin~K Lee, Kobbi Nissim, Sofya Raskhodnikova,
  and Adam Smith.
\newblock What can we learn privately?
\newblock {\em SIAM Journal on Computing}, 40(3):793--826, 2011.

\bibitem{wei2020federated}
Kang Wei, Jun Li, Ming Ding, Chuan Ma, Howard~H Yang, Farhad Farokhi, Shi Jin,
  Tony~QS Quek, and H~Vincent Poor.
\newblock Federated learning with differential privacy: Algorithms and
  performance analysis.
\newblock {\em IEEE Transactions on Information Forensics and Security},
  15:3454--3469, 2020.

\bibitem{zhao2020local}
Yang Zhao, Jun Zhao, Mengmeng Yang, Teng Wang, Ning Wang, Lingjuan Lyu, Dusit
  Niyato, and Kwok-Yan Lam.
\newblock Local differential privacy-based federated learning for internet of
  things.
\newblock {\em IEEE Internet of Things Journal}, 8(11):8836--8853, 2020.

\bibitem{liu2016paradrop}
Peng Liu, Dale Willis, and Suman Banerjee.
\newblock Paradrop: Enabling lightweight multi-tenancy at the network’s
  extreme edge.
\newblock In {\em 2016 IEEE/ACM Symposium on Edge Computing (SEC)}, pages
  1--13. IEEE, 2016.

\bibitem{marti2006taxonomy}
Sergio Marti and Hector Garcia-Molina.
\newblock Taxonomy of trust: Categorizing p2p reputation systems.
\newblock {\em Computer Networks}, 50(4):472--484, 2006.

\bibitem{shokri2015privacy}
Reza Shokri and Vitaly Shmatikov.
\newblock Privacy-preserving deep learning.
\newblock In {\em Proceedings of the 22nd ACM SIGSAC conference on computer and
  communications security}, pages 1310--1321, 2015.

\bibitem{wang2019beyond}
Zhibo Wang, Mengkai Song, Zhifei Zhang, Yang Song, Qian Wang, and Hairong Qi.
\newblock Beyond inferring class representatives: User-level privacy leakage
  from federated learning.
\newblock In {\em IEEE INFOCOM 2019-IEEE Conference on Computer
  Communications}, pages 2512--2520. IEEE, 2019.

\bibitem{bhowmick2018protection}
Abhishek Bhowmick, John Duchi, Julien Freudiger, Gaurav Kapoor, and Ryan
  Rogers.
\newblock Protection against reconstruction and its applications in private
  federated learning.
\newblock {\em arXiv preprint arXiv:1812.00984}, 2018.

\bibitem{li2018federated}
Tian Li, Anit~Kumar Sahu, Manzil Zaheer, Maziar Sanjabi, Ameet Talwalkar, and
  Virginia Smith.
\newblock Federated optimization in heterogeneous networks.
\newblock {\em arXiv preprint arXiv:1812.06127}, 2018.

\bibitem{rajput2019detox}
Shashank Rajput, Hongyi Wang, Zachary Charles, and Dimitris Papailiopoulos.
\newblock Detox: A redundancy-based framework for faster and more robust
  gradient aggregation.
\newblock In {\em Advances in Neural Information Processing Systems}, pages
  10320--10330, 2019.

\bibitem{chen2018draco}
Lingjiao Chen, Hongyi Wang, Zachary Charles, and Dimitris Papailiopoulos.
\newblock Draco: Byzantine-resilient distributed training via redundant
  gradients.
\newblock {\em arXiv preprint arXiv:1803.09877}, 2018.

\bibitem{zhao2018federated}
Yue Zhao, Meng Li, Liangzhen Lai, Naveen Suda, Damon Civin, and Vikas Chandra.
\newblock Federated learning with non-iid data.
\newblock {\em arXiv preprint arXiv:1806.00582}, 2018.

\bibitem{geng2015staircase}
Quan Geng, Peter Kairouz, Sewoong Oh, and Pramod Viswanath.
\newblock The staircase mechanism in differential privacy.
\newblock {\em IEEE Journal of Selected Topics in Signal Processing},
  9(7):1176--1184, 2015.

\bibitem{balle2018improving}
Borja Balle and Yu-Xiang Wang.
\newblock Improving the gaussian mechanism for differential privacy: Analytical
  calibration and optimal denoising.
\newblock In {\em International Conference on Machine Learning}, pages
  394--403. PMLR, 2018.

\bibitem{jayaraman2019evaluating}
Bargav Jayaraman and David Evans.
\newblock Evaluating differentially private machine learning in practice.
\newblock In {\em 28th $\{$USENIX$\}$ Security Symposium ($\{$USENIX$\}$
  Security 19)}, pages 1895--1912, 2019.

\bibitem{zhao2020privacy-utility}
Benjamin Zi~Hao Zhao, Mohamed~Ali Kaafar, and Nicolas Kourtellis.
\newblock Not one but many tradeoffs: Privacy vs. utility in differentially
  private machine learning.
\newblock In {\em Cloud Computing Security Workshop}. ACM CCS, 2020.

\bibitem{gilad2016cryptonets}
Ran Gilad-Bachrach, Nathan Dowlin, Kim Laine, Kristin Lauter, Michael Naehrig,
  and John Wernsing.
\newblock Cryptonets: Applying neural networks to encrypted data with high
  throughput and accuracy.
\newblock In {\em International Conference on Machine Learning}, pages
  201--210, 2016.

\bibitem{wagh2019securenn}
Sameer Wagh, Divya Gupta, and Nishanth Chandran.
\newblock Securenn: 3-party secure computation for neural network training.
\newblock {\em Proceedings on Privacy Enhancing Technologies}, 2019(3):26--49,
  2019.

\bibitem{zheng2019helen}
Wenting Zheng, Raluca~Ada Popa, Joseph~E Gonzalez, and Ion Stoica.
\newblock Helen: Maliciously secure coopetitive learning for linear models.
\newblock In {\em 2019 IEEE Symposium on Security and Privacy (SP)}, pages
  724--738. IEEE, 2019.

\bibitem{mishra2020delphi}
Pratyush Mishra, Ryan Lehmkuhl, Akshayaram Srinivasan, Wenting Zheng, and
  Raluca~Ada Popa.
\newblock Delphi: A cryptographic inference service for neural networks.
\newblock In {\em 29th $\{$USENIX$\}$ Security Symposium ($\{$USENIX$\}$
  Security 20)}, 2020.

\bibitem{mcmahan2017learning}
H~Brendan McMahan, Daniel Ramage, Kunal Talwar, and Li~Zhang.
\newblock Learning differentially private language models without losing
  accuracy.
\newblock {\em arXiv preprint arXiv:1710.06963}, 2017.

\bibitem{bonawitz2016practical}
Keith Bonawitz, Vladimir Ivanov, Ben Kreuter, Antonio Marcedone, H~Brendan
  McMahan, Sarvar Patel, Daniel Ramage, Aaron Segal, and Karn Seth.
\newblock Practical secure aggregation for federated learning on user-held
  data.
\newblock {\em arXiv preprint arXiv:1611.04482}, 2016.

\bibitem{bonawitz2017practical}
Keith Bonawitz, Vladimir Ivanov, Ben Kreuter, Antonio Marcedone, H~Brendan
  McMahan, Sarvar Patel, Daniel Ramage, Aaron Segal, and Karn Seth.
\newblock Practical secure aggregation for privacy-preserving machine learning.
\newblock In {\em Proceedings of the 2017 ACM SIGSAC Conference on Computer and
  Communications Security}, pages 1175--1191, 2017.

\bibitem{truex2019hybrid}
Stacey Truex, Nathalie Baracaldo, Ali Anwar, Thomas Steinke, Heiko Ludwig, Rui
  Zhang, and Yi~Zhou.
\newblock A hybrid approach to privacy-preserving federated learning.
\newblock In {\em Proceedings of the 12th ACM Workshop on Artificial
  Intelligence and Security}, pages 1--11, 2019.

\bibitem{truex2020ldp}
Stacey Truex, Ling Liu, Ka-Ho Chow, Mehmet~Emre Gursoy, and Wenqi Wei.
\newblock Ldp-fed: federated learning with local differential privacy.
\newblock In {\em Proceedings of the Third ACM International Workshop on Edge
  Systems, Analytics and Networking}, pages 61--66, 2020.

\bibitem{abad2020hierarchical}
M~Salehi~Heydar Abad, Emre Ozfatura, Deniz Gunduz, and Ozgur Ercetin.
\newblock Hierarchical federated learning across heterogeneous cellular
  networks.
\newblock In {\em ICASSP 2020-2020 IEEE International Conference on Acoustics,
  Speech and Signal Processing (ICASSP)}, pages 8866--8870. IEEE, 2020.

\bibitem{yuan2020hierarchical}
Jinliang Yuan, Mengwei Xu, Xiao Ma, Ao~Zhou, Xuanzhe Liu, and Shangguang Wang.
\newblock Hierarchical federated learning through lan-wan orchestration.
\newblock {\em arXiv preprint arXiv:2010.11612}, 2020.

\bibitem{briggs2020federated}
Christopher Briggs, Zhong Fan, and Peter Andras.
\newblock Federated learning with hierarchical clustering of local updates to
  improve training on non-iid data.
\newblock In {\em 2020 International Joint Conference on Neural Networks
  (IJCNN)}, pages 1--9. IEEE, 2020.

\bibitem{zhu2020deep}
Ligeng Zhu and Song Han.
\newblock Deep leakage from gradients.
\newblock In {\em Federated Learning}, pages 17--31. Springer, 2020.

\bibitem{petrek2001large}
Jozef Petrek and Volker Sledt.
\newblock A large hierarchical network star—star topology design algorithm.
\newblock {\em European transactions on telecommunications}, 12(6):511--522,
  2001.

\bibitem{messous2017computation}
Mohamed-Ayoub Messous, Hichem Sedjelmaci, Noureddin Houari, and Sidi-Mohammed
  Senouci.
\newblock Computation offloading game for an uav network in mobile edge
  computing.
\newblock In {\em 2017 IEEE International Conference on Communications (ICC)},
  pages 1--6. IEEE, 2017.

\bibitem{gueta2019sbft}
Guy~Golan Gueta, Ittai Abraham, Shelly Grossman, Dahlia Malkhi, Benny Pinkas,
  Michael Reiter, Dragos-Adrian Seredinschi, Orr Tamir, and Alin Tomescu.
\newblock Sbft: a scalable and decentralized trust infrastructure.
\newblock In {\em 2019 49th Annual IEEE/IFIP international conference on
  dependable systems and networks (DSN)}, pages 568--580. IEEE, 2019.

\bibitem{setty2012making}
Srinath~TV Setty, Richard McPherson, Andrew~J Blumberg, and Michael Walfish.
\newblock Making argument systems for outsourced computation practical
  (sometimes).
\newblock In {\em NDSS}, volume~1, page~17, 2012.

\bibitem{lua2005survey-p2poverlays}
Eng~Keong Lua, J.~Crowcroft, M.~Pias, R.~Sharma, and S.~Lim.
\newblock A survey and comparison of peer-to-peer overlay network schemes.
\newblock {\em IEEE Communications Surveys \& Tutorials}, 7(2):72--93, 2005.

\bibitem{lo2005super-peer-selection1}
Virginia Lo, Dayi Zhou, Yuhong Liu, Chris GauthierDickey, and Jun Li.
\newblock Scalable supernode selection in peer-to-peer overlay networks.
\newblock In {\em 2nd International Workshop on Hot Topics in Peer-to-Peer
  Systems}, pages 18--25, 2005.

\bibitem{mahdy2007super-peer-selection2}
Ahmed~M. Mahdy, Jitender~S. Deogun, and Jun Wang.
\newblock A dynamic approach for the selection of super peers in ad hoc
  networks.
\newblock In {\em Sixth International Conference on Networking {(ICN})}. {IEEE}
  Computer Society, 2007.

\bibitem{li2019super-peer-selection3}
Li~Qing, Fu~Xuan-li, Hou Yu-ke, and He~Wan-jie.
\newblock Super-peer selection algorithm based on ahp in mobile peer-to-peer
  network.
\newblock In {\em 7th International Conference on Information Technology: IoT
  and Smart City}, ICIT, page 305–309, New York, NY, USA, 2019. Association
  for Computing Machinery.

\bibitem{castro2002practical}
Miguel Castro and Barbara Liskov.
\newblock Practical byzantine fault tolerance and proactive recovery.
\newblock {\em ACM Transactions on Computer Systems (TOCS)}, 20(4):398--461,
  2002.

\bibitem{awerbuch1985complexity}
Baruch Awerbuch.
\newblock Complexity of network synchronization.
\newblock {\em Journal of the ACM (JACM)}, 32(4):804--823, 1985.

\bibitem{arvind1994probabilistic}
Koshal Arvind.
\newblock Probabilistic clock synchronization in distributed systems.
\newblock {\em IEEE Transactions on Parallel and Distributed Systems},
  5(5):474--487, 1994.

\bibitem{silberschatz1979communication}
Abraham Silberschatz.
\newblock Communication and synchronization in distributed systems.
\newblock {\em IEEE Transactions on Software Engineering}, pages 542--546,
  1979.

\bibitem{balle2018privacy}
Borja Balle, Gilles Barthe, and Marco Gaboardi.
\newblock Privacy amplification by subsampling: Tight analyses via couplings
  and divergences.
\newblock {\em arXiv preprint arXiv:1807.01647}, 2018.

\bibitem{balle2020privacy}
Borja Balle, Peter Kairouz, H~Brendan McMahan, Om~Thakkar, and Abhradeep
  Thakurta.
\newblock Privacy amplification via random check-ins.
\newblock {\em arXiv preprint arXiv:2007.06605}, 2020.

\bibitem{bittau2017prochlo}
Andrea Bittau, {\'U}lfar Erlingsson, Petros Maniatis, Ilya Mironov, Ananth
  Raghunathan, David Lie, Mitch Rudominer, Ushasree Kode, Julien Tinnes, and
  Bernhard Seefeld.
\newblock Prochlo: Strong privacy for analytics in the crowd.
\newblock In {\em Proceedings of the 26th Symposium on Operating Systems
  Principles}, pages 441--459, 2017.

\bibitem{erlingsson2019amplification}
{\'U}lfar Erlingsson, Vitaly Feldman, Ilya Mironov, Ananth Raghunathan, Kunal
  Talwar, and Abhradeep Thakurta.
\newblock Amplification by shuffling: From local to central differential
  privacy via anonymity.
\newblock In {\em Proceedings of the Thirtieth Annual ACM-SIAM Symposium on
  Discrete Algorithms}, pages 2468--2479. SIAM, 2019.

\bibitem{feldman2022hiding}
Vitaly Feldman, Audra McMillan, and Kunal Talwar.
\newblock Hiding among the clones: A simple and nearly optimal analysis of
  privacy amplification by shuffling.
\newblock In {\em 2021 IEEE 62nd Annual Symposium on Foundations of Computer
  Science (FOCS)}, pages 954--964. IEEE, 2022.

\bibitem{cohen2017emnist}
Gregory Cohen, Saeed Afshar, Jonathan Tapson, and Andre Van~Schaik.
\newblock Emnist: Extending mnist to handwritten letters.
\newblock In {\em 2017 International Joint Conference on Neural Networks
  (IJCNN)}, pages 2921--2926. IEEE, 2017.

\bibitem{cifar10}
Alex Krizhevsky, Vinod Nair, and Geoffrey Hinton.
\newblock Cifar-10 (canadian institute for advanced research).

\bibitem{cifar100}
Alex Krizhevsky, Vinod Nair, and Geoffrey Hinton.
\newblock Cifar-100 (canadian institute for advanced research).

\bibitem{DBLP:journals/corr/abs-2102-05975}
Marlotte Pannekoek and Giacomo Spigler.
\newblock Investigating trade-offs in utility, fairness and differential
  privacy in neural networks.
\newblock {\em CoRR}, abs/2102.05975, 2021.

\bibitem{zhang2018unreasonable}
Richard Zhang, Phillip Isola, Alexei~A Efros, Eli Shechtman, and Oliver Wang.
\newblock The unreasonable effectiveness of deep features as a perceptual
  metric.
\newblock In {\em Proceedings of the IEEE conference on computer vision and
  pattern recognition}, pages 586--595, 2018.

\end{thebibliography}

\end{document}